\documentclass{article}
\usepackage{graphicx} 
\usepackage{hyperref}
\usepackage{authblk}

\usepackage[
backend=biber,
style=numeric,
sorting=ynt
]{biblatex}
\title{\textbf{Addressing the Abstraction and Reasoning Corpus via Procedural Example Generation}}
\author{Michael Hodel}
\date{
    Institute of Neuroinformatics, University of Zurich and ETH Zurich
}

\begin{document}

\maketitle

\abstract
This work presents code to procedurally generate examples for the ARC training tasks \cite{otmoi} \cite{arc}. For each of the 400 tasks, an example generator following the transformation logic of the original examples was created. In effect, the assumed underlying distribution of examples for any given task was reverse engineered by implementing a means to sample from it. An attempt was made to cover an as large as reasonable space of possible examples for each task. That is, whenever the original examples of a given task may be limited in their diversity e.g. by having the dimensions of the grids, the set of symbols or number of objects constant or within tight bounds, even though the transformation does not require it, such constraints were lifted. Having access to not just a few examples per task, as the case for ARC, but instead very many, should enable a wide range of experiments that may be important stepping stones towards making leaps on the benchmark \footnote{The code is available at \href{https://github.com/michaelhodel/re-arc}{https://github.com/michaelhodel/re-arc}}.

\section{Motivation}
The Abstraction and Reasoning Corpus (ARC) is a dataset intended to serve as a benchmark for general intelligence. It consists of 1000 unique and diverse tasks asking for the test-taker to, given a number of demonstration pairs of input and output grids, predict the output grid for test input grids. The difficulty of ARC for machine learning approaches is largely a consequence of the great diversity of tasks as well its few-shot nature. Even after more than four years since its publication, ARC remains unsolved. While many attempts have been made to solve ARC, what seems generally lacking are fundamental scientific experiments. The data generation process in this work attempts to enable such experiments for addressing the latter of those two aspects, namely sample-efficient learning. A simple such experiment would be comparing various model architectures and learning algorithms in a setting of having a separate model trained for each task: For a given fixed number of examples, how well can the model generalize to an unseen set of examples of the same task, on average across all the training tasks? There are many opportunities in this realm of within-task generalization, e.g. addressing the question of how the sample-efficiency of learning algorithms might be improved by building a curriculum that increases example difficulty over the course of training - as opposed to training on instances of the full range of difficulties throughout the entire training - or by other means of constructing a curriculum that is not fully random or uniform in terms of some metric. While it seems implausible that strong statements about suitability of approaches trying to solve ARC could be made directly based on results of such experiments, they still may a means to gather important insights into how progress on ARC could be made.

\section{Task Generalization}
Broad generalization of the tasks and therefore large sample spaces were attempted. Generating new examples for an arbitrary ARC task via say a set of fixed augmentation schemes would yield only minimal diversity and, more importantly, can't work in general in the first place, as the augmentations may break examples. Hence, here, new examples are constructed from the ground up instead. Each value for all relevant degrees of freedom like grid dimension or number of objects, where applicable, is determined randomly and individually for each example, e.g. sampling integers determining the grid dimensions or number of symbols or objects, as well as generating randomly shaped and colored objects, or randomly placing objects on a grid, etc. It was generally attempted that each value within the generation process that need not be hardcoded was sampled instead. Further, it was attempted that each two components that do not logically depend upon each other are sampled separately: E.g. for most tasks, the grid height and width are independent from each other. Lastly, an effort was made for each value of any degree of freedom to be sampled from an as large a set as reasonably possible. For example, if a task features an objectness notion and the number of objects may vary freely, then the generator should allow for examples featuring only a single or also very many objects.

There are some limitations to the notion of broadest-most possible generalization however, and the realized sample space does not always cover the actual space of possible examples. These limitations are a direct result of favouring simpler or faster code over covering certain edge cases or fully exhausting other pockets of the theoretical ideal sample space, but also as a result of attempting to retain each ARC task's spirit. Still, each generator is capable of generating at least 10'000 unique examples, and the vast majority of the generators even have a vastly greater sample space.

Even if the generators could have been written to allow arbitrarily large grids or number of symbols, to stay at least somewhat close to the original ARC format, the number of symbols was kept at 10 and the maximum grid height and width were kept at 30. Nevertheless, there are three obvious ways in which the generated data fundamentally differ from the ARC data: There are many more examples per task, each task is typically much more general, i.e. the diversity of examples within a task is greater, and examples may be either much simpler or also significantly harder, e.g. since they may require dealing with many more elements.

\section{Example Generation}
Each generator is a standalone Python function merely making use of the DSL and functions from the random module from the standard library. The median generator consists of 40 lines of code and uses 22 DSL primitive calls and 10 calls to the random module \cite{dsl}. Even though there is a great degree of overlap in the utilized techniques across the generators, no separate utilities were created, apart from a wrapper around the random function for controlling example difficulty. The interface of every generator is simple, in that, when called, it returns a single example. Some of the strategies used are briefly presented here, to give some insight into how the generation processes were implemented.

As the notion of objects is crucial to many ARC tasks, generating and placing objects was frequently required. Typically, to generate a random object, first a size of the object in terms of number of pixels was sampled, and then - conditional on some notion of adjacency and maximum height and width of the object - neighboring pixel were iteratively added to the object, and then the object was colored. To place objects on grids, commonly one of two strategies was employed. The first is rejection sampling, i.e. selecting a still unoccupied location and verifying that placing the object there does not result in unintended occlusions. Alternatively, e.g. in cases where rejection sampling was expected to give too high failure rates, object locations were determined by sampling from a pre-filtered set of candidate locations that are guaranteed to work.

Oftentimes, where possible, the requirement for a certain specific orientation - e.g. with respect to rotation or mirroring - was abstracted away, and to keep the generation process simple, variation in the orientation was simply introduced by applying respective transformation functions to both the input and output grid at the end of the generation process - as opposed to accounting for the possible orientations already during the process of drawing onto grids, which may unnecessarily complicate the implementation without adding functionality.

Furthermore, for some tasks, the underlying transformation is an invertible function, and for some of those, the inverse is more straight-forward to implement. For such tasks, as opposed to sampling some input grid and then applying some transformation to it - or, as was very frequently done, constructing both the input grid in tandem - the output grid was usually generated first and then the inverse transformation was applied to arrive at the input grid.

\section{Example Verification}
It was deemed crucial to have some means of ensuring a high quality of the generated examples. However, assessing the quality of the generated examples manually is in general highly infeasible due to their potential numerosity as well as the human tendency to make mistakes, especially careless mistakes for more involved examples. Therefore, each generator goes alongside a corresponding verifier. A verifier is simply a task-specific function that transforms any input grid that is valid for that task to its correct corresponding output grid. Hence, the verifiers can be used to keep only generated examples on which the verifier works, i.e. which are valid. Each verifier also works on all the original examples of the corresponding task \footnote{There are a few exceptions however: For the tasks a8d7556c, 6cf79266, 469497ad, 9edfc990 and e5062a87, either the original examples contained bugs or the task logic has been slightly adapted, and for the tasks 97a05b5b, 4290ef0e, 7e0986d6, 53b68214, 29ec7d0e and a64e4611 there is each one example in the original task for which the verifier fails.}. The verifiers oftentimes directly build on the solver programs for the original ARC tasks as provided with the DSL. The verifiers are all written entirely within the ARC-DSL. The median verifier program consists of 18 DSL primitive calls.

\section{Example Difficulty}
While novelty is an important factor co-determining difficulty, it is not directly considered during generation, as it is context-dependent. Instead, an emphasis is put on cardinalities, a more explicit proxy measure for difficulty: Arguably, examples involving more elements to deal with (rows, pixels, symbols, objects, etc.) tend to be more difficult. Each generator allows to pass parameters to control the example difficulty in this way. Example difficulty here is defined by an interval within the range [0, 1], which controls the uniform random number generation used for sampling cardinalities. For example, if an input grid may be of any height 1 through 30, and bounds [1/3, 1] are passed as parameters, then any height in range [10, 30] may be sampled; or if there may be between two and five objects and bounds [0, 1/2] are used, then there will be either two or three objects. The interval is used in any place where sampling larger values generally corresponds to greater difficulty, and may include anything from grid height, width, number of symbols, number of objects, sizes of objects, number of symbols per objects, distances to travel, etc. Hence, intervals closer to 1 and further from 0 tend to result in more difficult examples. This sampling is achieved via a simple wrapper function around the random function of the random module. Robustly controlling example difficulty was often nontrivial. For instance, setting the upper bound for the number of objects too high may lead to extremely inefficient object sampling or placement given a large upper bound for the difficulty. Also, there were trade-offs, e.g. between the number of objects and their sizes: both individually arguably increase example difficulty, but having both at their respective maximum values is impossible.

The main motivation for allowing this somewhat explicit control over example difficulty bounds was to enable within-task generalization capabilities of models: \textit{Can a model also solve other (more difficult) examples from the same task?} This generalization capability is not to be confused with the much more relevant notion of across-task generalization: \textit{Can a model also solve different (but related) tasks?} Controlling for example difficulty via the means of these parameters could be achieved by generating levels, in the simplest case two sets of examples, the easier of which uses interval [0, 0.5], and the harder of which uses interval [0.5, 1]. However, creating too many levels of difficulty may not always work out nicely for each task, since the ranges of possible values may be too small and would render the pruned range empty. An issue with creating too many levels is that in many cases, some degrees of freedom are not independent: E.g. if the grid is smaller, then the upper bound for the number of objects placeable is also smaller. For this reason, too small ranges per level would hence not allow generation of certain instances in the first place. The aim was to not over-engineer things while still retaining a uniform interface and large sample space for all tasks.

While this means to control example difficulty could allow for tailored datasets or curricula building during the generation process, it is, for the reasons explained above, brittle to some degree. Hence, it is suggested - depending on the experimental setup - to also or instead consider the alternative of controlling for example difficulty in a post-hoc manner, where examples are filtered or ordered after the generation, based on some other metrics. Two metrics of difficulty, each in range $[0, 1]$, are provided: Firstly, the \textit{RNG-Difficulty} defined as the average of the numbers sampled from the provided bounds in the process of generating an example, is heavily flawed for a number of reasons. For example, processes may fail and the numbers sampled there still influence this metric. Also, the random number influencing the number of objects has the same weight as e.g. the random number influencing the number of pixels of an individual object have the same weight. Secondly, the \textit{PSO-Difficulty} is defined as $(P/1800 + S/10 + O/P) / 3$, where P is the number of pixels, S the number of symbols and O the number of objects, and an object is defined as a set of pixels of the same symbol that are all horizontally or vertically connected. Note that these metrics are mainly intended to serve as inspiration, the metrics used for defining grouping or ordering of data should be tailored to the questions that the experiments at hand try to address.

For about a third of the tasks, the number of symbols that occur in each example is fixed. Over half the tasks however have at least 6 different values for number of symbols an example may take on. Such tasks could be used to conduct experiments for investigating the capabilities of models to generalize to examples containing $n$ symbols, after having been trained on examples containing less than $n$ symbols. Furthermore, almost all of the tasks have at least 10 different values for the number of pixels an example may take on, and the vast majority of tasks have at least 10 different values for the number of objects an example may take on (using a simplistic uniform objectness notion of treating each each horizontally or vertically connected set of pixels of a single color as an object), hence experiments for assessing generalization capabilities along those degrees of freedoms may also be feasible for respectively adequate subsets of tasks.

\section{Limitations}
Note that the \textit{true underlying transformation logic} as arguably the only or at least by far most crucial defining characteristic of a task is not well-defined. The overall aim was to have the widest possible interpretation of the transformation logic for any given task, such that most individuals only familiar with the original examples could solve any generated example within three guesses. However, this is not guaranteed. The generalized task is simply \textit{defined} as the set of pairs of input and output grids producible by the generator for which the verifier function, when applied to the input grid, returns the output grid. Importantly, this does not mean that the verifier would work for every conceivable example a human might associate with that task, nor every example producible by the generator. Moreover, it does not imply that every example that works with the verifier can necessarily be produced by the generator, nor does it ensure that the generated examples are solvable by humans. Still, providing verifiers should add a lot of credibility to the data generation process.

While for most generators, all or a vast majority of producible examples are valid, there are some with relatively low sample-efficiency in terms of the percentage of generated examples that are valid. Generation of new verified examples for a given task may also be slow for various other reasons, such as a slow generation process itself, a small sample space resulting in more frequent duplicate examples, or slow verification. Still, approximately 1'000 verified unique examples per second can be generated for the median task.

\medskip

\printbibliography

\end{document}